%% file: main.tex
\documentclass[10pt,twocolumn,letterpaper]{article}

\usepackage{cvpr}              
\usepackage[accsupp]{axessibility}
\usepackage{times}
\usepackage{epsfig}
\usepackage{graphicx}
\usepackage{amsmath}
\usepackage{amssymb}
\usepackage[numbers,sort&compress]{natbib}
\usepackage{color}
\usepackage{multirow}
\usepackage{booktabs}
\usepackage{soul}
\usepackage{colortbl}
\usepackage[ruled,linesnumbered]{algorithm2e}
\definecolor{mygray}{gray}{.9}
\definecolor{mypink}{rgb}{.99,.91,.95}
\definecolor{mycyan}{cmyk}{.3,0,0,0}
\def\ie{\emph{i.e.}}
\def\eg{\emph{e.g.}}
\def\etal{{\em et al.}}
 
\input{preamble}

\definecolor{cvprblue}{rgb}{0.21,0.49,0.74}
\usepackage[pagebackref,breaklinks,colorlinks,citecolor=cvprblue]{hyperref}

\def\paperID{1704} 
\def\confName{CVPR}
\def\confYear{2024}

\title{Genuine Knowledge from Practice: \\Diffusion Test-Time Adaptation for Video Adverse Weather Removal}

\author{
Yijun Yang$^{1}$, Hongtao Wu$^{1}$,  Angelica I. Aviles-Rivero$^{2}$, Yulun Zhang$^{3}$, Jing Qin$^{4}$,  Lei Zhu$^{1,5}$\thanks{Lei Zhu (leizhu@ust.hk) is the corresponding author.}\\
\footnotesize $^{1}$The Hong Kong University of Science and Technology (Guangzhou)\quad
\footnotesize $^{2}$University of Cambridge\quad $^{3}$ETH Zürich\\ \vspace{-0.5mm}
\footnotesize $^{4}$The Hong Kong Polytechnic University\quad
\footnotesize $^{5}$The Hong Kong University of Science and Technology\\
{\tt\small Project page: \url{https://github.com/scott-yjyang/DiffTTA}}}

\begin{document}
\maketitle
\input{section0-abstract.tex}

\input{section1-introduction}
\input{section2-related}
\input{section3-method}
\input{section4-experiments}

\input{section5-conclusion}


{
    \small
    \bibliographystyle{ieeenat_fullname}
    \bibliography{main}
}

\clearpage
\input{X_suppl}

\end{document}

%% file: preamble.tex
\usepackage[dvipsnames]{xcolor}


%% file: section0-abstract.tex
\begin{abstract}
Real-world vision tasks frequently suffer from the appearance of unexpected adverse weather conditions, including rain, haze, snow, and raindrops.
In the last decade, convolutional neural networks and vision transformers have yielded outstanding results in single-weather video removal. However, due to the absence of appropriate adaptation, most of them fail to generalize to other weather conditions. Although ViWS-Net is proposed to remove adverse weather conditions in videos with a single set of pre-trained weights, it is seriously blinded by seen weather at train-time and degenerates when coming to unseen weather during test-time.
In this work, we introduce test-time adaptation into adverse weather removal in videos, and propose the first framework that integrates test-time adaptation into the iterative diffusion reverse process.
Specifically, we devise a diffusion-based network with a novel temporal noise model to efficiently explore frame-correlated information in degraded video clips at training stage. During inference stage, we introduce a proxy task named Diffusion Tubelet Self-Calibration to learn the primer distribution of test video stream and optimize the model by approximating the temporal noise model for online adaptation.
Experimental results, on benchmark datasets, demonstrate that our Test-Time Adaptation method with Diffusion-based network(Diff-TTA) outperforms state-of-the-art methods in terms of restoring videos degraded by seen weather conditions. Its generalizable capability is also validated with unseen weather conditions in both synthesized and real-world videos. 

\end{abstract}

%% file: section1-introduction.tex
\section{Introduction}

\begin{figure}
\centering
\includegraphics[width=\columnwidth]{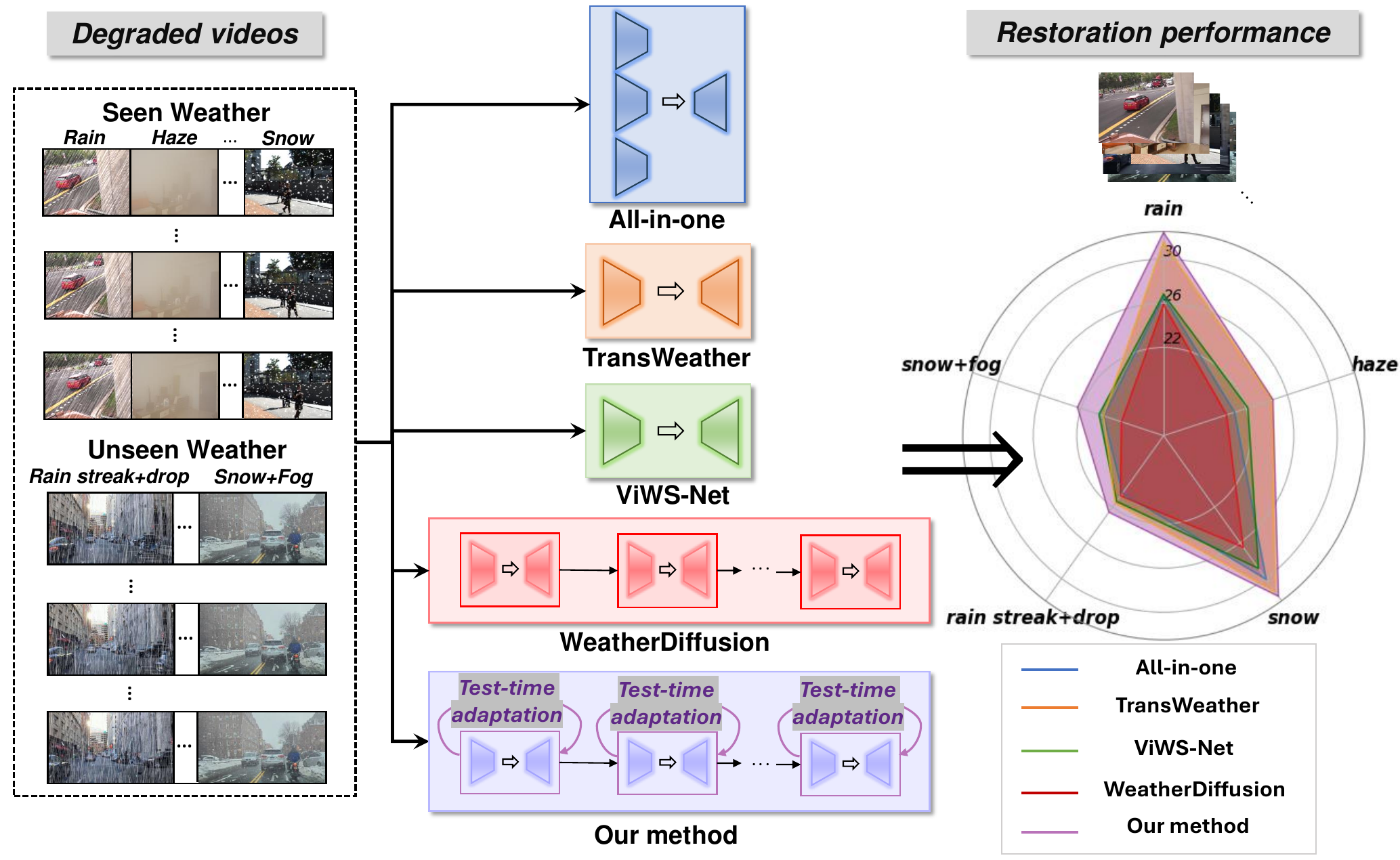}
\caption{\textbf{Overview of the existing all-in-one adverse weather removal methods.} Our approach can achieve superior performance not only in seen weather conditions but also in unseen weather conditions with a single set of pre-trained weights by diffusion test-time adaptation. In particular, our Diff-TTA is 90$\times$ more efficient than WeatherDiffusion.}
\label{fig:overview}
\vspace{-5mm}
\end{figure}

Adverse weather like rain, snow, and haze is very common in many existing outdoor videos. The presence of adverse weather usually decreases the visibility of the captured videos and significantly impairs the performance of subsequent high-level vision applications such as object detection, semantic segmentation and autonomous driving. To mitigate these negative effects, Convolutional Neural Networks (CNNs) and Transformers are frequently utilized in the literature. Also, Diffusion models have emerged as the de-facto generative model for image synthesis~\cite{ho2020denoising,dhariwal2021diffusion,singh2022high} and have been developed for single-image adverse weather removal~\cite{ozdenizci2023restoring}. 
While deep neural networks can achieve decent results on test points within the distribution, real-world applications often demand robustness. This is especially crucial when facing unknown input perturbations, changes in weather conditions, or other sources of distribution shift. 
Existing single-weather removal approaches~\cite{li2018video,zhang2021learning,chen2023snow} require follow-up training to adapt to specific domains. This leads to the switching between different parameter sets, which can make the pipeline cumbersome.

\begin{figure}
\centering
\includegraphics[width=\columnwidth]{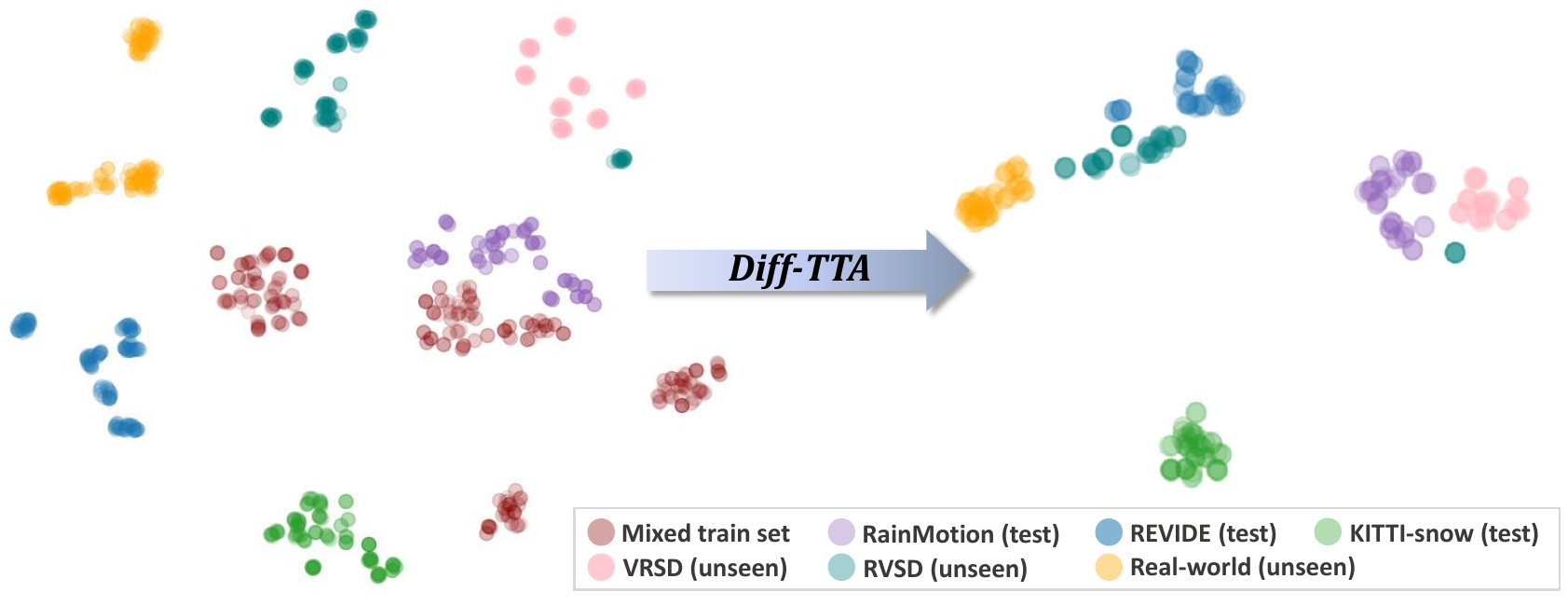}
\caption{\textbf{Our Diff-TTA enables weather removal models to overcome unseen weather corruptions.} We use t-SNE~\cite{van2008visualizing} to visualize features from the last feature extractor layer of each dataset. Obviously, unseen data points tend to approximate the seen ones after adaptation, which means Diff-TTA can categorize unknown degradation into known distribution. (`Real-world' contains video clips simultaneously degraded by fog and snow.)}
\label{fig:tsne}
\vspace{-5mm}
\end{figure}

Recently, all-in-one adverse weather removal has attracted the attention of the community, which aims to handle diverse weather conditions with a generic model~\cite{li2020all,valanarasu2022transweather,chen2022learning,yang2023video}. 
However, the potential issue of frequently failing to recover from out-of-distribution conditions remains largely unaddressed due to the finiteness of training data and the absence of a test data adaptation mechanism.
As shown in Figure~\ref{fig:overview}, generic models suffer significant degradation when handling all the possible weather conditions beyond the scope of their training data.
Furthermore, a common aspect among many real-world applications is the necessity for online adaptation with limited data. This is particularly relevant in scenarios such as autonomous vehicles or drones, where a vision model must continually process a video stream and adjust its predictions. 
To achieve effective adaptation, it is crucial to leverage the incoming data in a way that is beneficial and does not introduce harmful biases, even when there are distribution shifts between the training and test data.
\textit{Thus, akin to how humans update knowledge through practice, the task of removing any weather condition through online adaptation using a single set of pre-trained weights is both pragmatic and promising, albeit challenging.}

In this paper, we construct the diffusion-based framework for video adverse weather removal.
Yet, diffusion models usually entail significant training overhead, hindering the technique’s broader adoption in the research community of video tasks.
Instead of introducing massive parameters for temporal encoding, we design an efficient strategy based on DDIM~\cite{song2020denoising} to preserve temporal correlation, specifically \textit{temporal noise model}, in order to alleviate its training load. 
In contrast to the vanilla noise model, which independently samples noise for each frame from a standard Gaussian distribution, our temporal noise model takes into account the inter-frame relationship via time series models. 
%
\textit{More importantly, to promptly address unexpected weather conditions, we introduce test-time adaptation techniques by recalibrating the model's parameters using only the degraded test data. }
%
The mainstream of test-time adaptation methods~\cite{wang2020tent,iwasawa2021test,zhang2022memo} manipulate and iteratively update the normalization statistics of CNNs and Vision Transformer using specific proxy tasks at the test phase. However, these methods often fail when transferring to low-level tasks. 
Additionally, incorporating such iterative adaptation into these models significantly increases the cost of inference, multiplying it by the number of iterations, compared to the original counterpart. 

In contrast, we highlight the advantage of integrating the diffusion reverse process and test-time adaptation based on their common characteristics of iteration. As far as we know, there has been no work to emerge the possibility of Diffusion models and test-time adaptation.
Based on our diffusion-based framework, we design a proxy task and efficiently incorporate it into each denoising iteration when reversing our temporal diffusion process. 
Specifically, we develop a surrogate task called Diffusion Tubelet Self-Calibration (Diff-TSC), which is supervised by temporal noise model. This is designed to be task-agnostic and does not require any alterations to the training procedure. Rather than directly traversing all fine-grained details, we extract a few tubelets to taste the primer distribution of the frames degraded by unknown weather. This approach facilitates rapid optimization of the diffusion model while also reducing computational costs. 
As displayed in Figure~\ref{fig:tsne}, test-time adaptation helps to improve the model's robustness in real-world scenarios by mitigating the gap between seen and unseen weather corruptions. 

\noindent Our contributions can be summarized as:
\begin{itemize}
    \item We introduce the first diffusion-based framework for all-in-one adverse weather removal in videos. This framework efficiently leverages temporal redundancy information through the novel temporal noise model.
    \item To improve the model's robustness against unknown weather, we are the first to introduce test-time adaptation by incorporating a proxy task into the diffusion reverse process to learn the primer distribution of the test data. 
    \item We extensively evaluate our approach on seen weather (\eg, rain, haze, snow) and unseen weather (\eg, rain streak+raindrop, snow+fog). The superior performance over synthesized and real-world videos consistently validates its effectiveness and generalizable ability.
\end{itemize}

%% file: section2-related.tex
\section{Related Work}
\label{sec:relatedwork}

\textbf{All-in-one Adverse Weather Removal.} 
Over the past decade, The community of low-level vision has paid much attention to single weather removal, \eg, deraining~\cite{chen2018robust,li2018video,liu2018erase,wang2022rethinking,yan2021self,yang2019frame,yang2020self,yue2021semi}, dehazing~\cite{ren2018deep,zhang2021learning,liu2022phase,xu2023video}, desnowing~\cite{chen2020jstasr,zhang2021deep,chen2023snow,chen2023uncertainty,chen2023cplformer}, both on the image level and the video level.
%
%
Due to the limitations of existing models in adapting to different weather conditions, researchers have recently been investigating the use of a single model instance for multiple adverse weather removal tasks.
At the image level, following the pioneering work All-in-one~\cite{li2020all}, TransWeather~\cite{valanarasu2022transweather} and TKL~\cite{chen2022learning} were introduced to tackle many weather conditions with the single encoder and decoder.
Zhu \etal~\cite{zhu2023learning} elaborated a two-stage training strategy to explore the weather-general and weather-specific information for the sophisticated removal of diverse weather conditions.
Ye \etal~\cite{ye2023adverse} introduced high-quality codebook priors learned by a pre-trained VQGAN to restore texture details from adverse weather degradation.
The aforementioned methods failed to capture redundant information from temporal space. 
While they can be adapted for frame-by-frame adverse weather removal, incorporating temporal data in video-level methods provides a more robust solution for dealing with multiple adverse weather conditions in videos.
Very recently, Yang~\cite{yang2023video} introduced `ViWS-Net', the first video-level approach. This method uses weather messenger tokens to transfer weather knowledge across frames. It also employs adversarial learning to distinguish the weather type, thereby emphasizing weather-invariant information.
However, they did not consider online adaptation of the deep model, making the potential collapse for unseen scenarios degraded by unknown weather.

\textbf{Diffusion Models.}
Diffusion model is well-known as a generative modeling approach for its superior achievements in image synthesis and generation tasks~\cite{ho2020denoising,dhariwal2021diffusion,singh2022high,yang2023diffmic,zhang2023adding,liu2023more}.
In essence, Denoising Diffusion Probabilistic Models (DDPM)~\cite{ho2020denoising} adopted parameterized Markov chain to optimize the lower variational bound on the likelihood function, which can simulate a diffusion process to iteratively improve the quality of target distribution than other generative models. 
%
%
For restoration tasks, Luo \etal~\cite{luo2023image} applied stochastic differential equations featuring mean reversion in diffusion models to improve low-quality images.
DiffIR~\cite{xia2023diffir} provided a strong diffusion baseline for image restoration by estimating a compact IR prior representation.
Zhang \etal~\cite{zhang2023unified} proposed a unified conditional framework based on diffusion models for image restoration by multi-source guidance of a lightweight U-Net.
Özdenizci \etal~\cite{ozdenizci2023restoring} designed a patch-based diffusion modeling approach called WeatherDiffusion to make the network size-agnostic. 
While this technique utilizes diffusion models for image weather removal, its very slow inference speed restricts practical use.
Nevertheless, there is no work adapting diffusion models to \textit{video} adverse weather removal.

\textbf{Test-time Adaptation.}
Recently, there has been a growing interest in adapting deep models using exclusively sequential, unlabeled test data.
This contrasts with the traditional unsupervised domain adaptation, which asks the access to both training data and a sufficient amount of unlabeled test data. Test-time adaptation techniques allow for model updates using distributional statistics from either a single instance or batch of unlabeled test data.
Typically, TENT~\cite{wang2020tent} focuses on minimizing the model's prediction entropy on the target data. 
T3A~\cite{iwasawa2021test} adjusts the classifier of a trained source model by computing a pseudo-prototype representation of different classes using unlabeled test data.
MEMO~\cite{zhang2022memo} aims to minimize marginal entropy, compelling the deep model to consistently predict across various augmentations. This approach enforces the invariances of these augmentations and ensures prediction confidence.
Chen \etal~\cite{chen2022contrastive} proposed to apply self-supervised contrastive learning for target feature learning, which was trained jointly with pseudo labeling.
Although test-time adaptation methods are increasingly used in image classification~\cite{niu2022efficient,wang2022continual,boudiaf2022parameter,nguyen2023tipi,zhang2023domainadaptor} and segmentation~\cite{hu2021fully,liu2022single,yang2022dltta} tasks, their potential in adverse weather removal tasks remains unexplored. 
In our work, the nature of the iterative sampling process in Diffusion model significantly promotes the adaptation to the target distribution.
Video scenarios further allow for the accumulation of the adaptation knowledge for the subsequent frames.

%% file: section3-method.tex
\vspace{-1mm}
\section{Preliminaries: Diffusion Models}
The diffusion model is composed of a forward and a reverse process. The forward process is defined as a discrete Markov chain of length $T$: $q(x_{1:T}|x_0)=\prod_{t=1}^{T}q(x_t|x_{t-1})$. For each step $t\in[1,T]$ in the forward process, a diffusion model adds noise $\epsilon_t$ sampled from the Gaussian distribution $\mathcal{N}(0,\mathbf{I})$ to data $x_{t-1}$ and obtains disturbed data $x_t$ from $q(x_t|x_{t-1})=\mathcal{N}(x_t;\sqrt{1-\beta_t}x_{t-1},\beta_t\mathbf{I})$. $\beta_t$ decides the ratio of noise at timestep $t$.
Noticeably, instead of sampling sequentially along the Markov chain, we can sample $x_t$ at any time step $t$ in the closed form via $q(x_t|x_{0})=\mathcal{N}(x_t;\sqrt{\bar{\alpha}_t}x_{t-1},(1-\bar{\alpha}_t)\mathbf{I})$, where $\bar{\alpha}_t=\prod^t_{s=1}(1-\beta_s)$. To parameterize the Gaussian distribution, the neural network $\epsilon_\theta$ is introduced, which is optimized by the objective of DDPM~\cite{ho2020denoising}:
\begin{equation}
    \mathcal{L}_S = ||\epsilon-\epsilon_\theta(\sqrt{\bar{\alpha}_t}x_0+\sqrt{1-\bar{\alpha_t}}\epsilon, t)||^2_2 .
\label{eq:loss}
\end{equation}
In the reverse process, the diffusion model gradually denoises the randomly sampled Gaussian noise to the high-quality output $x_0$ through the predicted noise by the well-trained $\epsilon_\theta$. This process is also defined as a Markov chain: $p_\theta(x_{0:T})=p(x_T)\prod_{t=1}^Tp_\theta(x_{t-1}|x_t)$, and $p_\theta(x_{t-1}|x_t)=\mathcal{N}(x_{t-1};\mu_\theta(x_t,t),\sigma^2_t\mathbf{I})$. The mean and variance are $\mu_\theta(x_t,t)=\frac{1}{\sqrt{\alpha_t}}(x_t-\frac{\beta_t}{\sqrt{1-\bar{\alpha}_t}}\epsilon_\theta(x_t,t))$, $\sigma^2_t=\frac{1-\bar{\alpha}_{t-1}}{1-\bar{\alpha}_t}\beta_t$, respectively. 

\section{Method}
\label{sec:method}

\begin{figure*}[!t]
\centering
\includegraphics[width=0.95\textwidth]{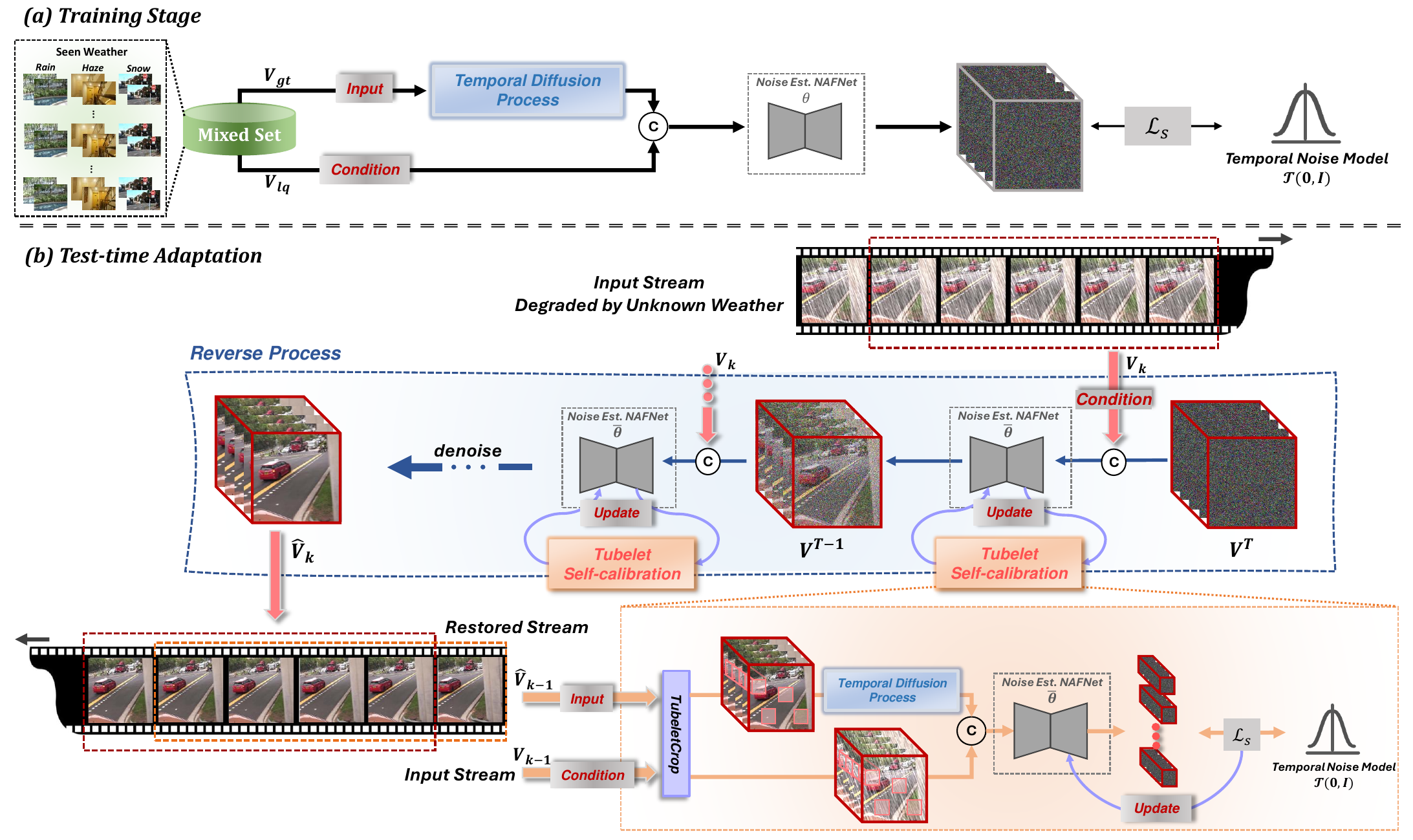}
\caption{\textbf{Overview of our Diff-TTA framework for video adverse weather removal.} In the training stage, through the temporal diffusion process, we randomly add temporal noise on the clean video clip $\mathbf{V}_{gt}$ from the mixed training set. Then, the denoising NAFNet $\epsilon_\theta$ is trained by estimating the applied noise, which is conditioned by the degraded counterpart $\mathbf{V}_{lq}$ by seen weather types. In the test stage, the proxy task, Diff-TSC, where the cropped tubelets of the last restored pair $\{\mathbf{V}_{k-1},\hat{\mathbf{V}}_{k-1}\}$ are utilized to learn the primer distribution, is incorporated into each timestep of denoising the randomly sampled temporal noise $\mathbf{V}^T_k$ for the iterative online adaptation.}
\label{fig:framework}
\vspace{-5mm}
\end{figure*}

In this work, we devise the first diffusion-based model to remove arbitrary adverse weather in videos with one set of model parameters. We follow an end-to-end formulation of video adverse weather removal, which reads:
\begin{equation}
    \begin{aligned} 
    \hat{\mathbf{V}}_k = \phi(\textbf{V}_k),
    \mathbf{V}_k = \{I_{i}\}_{i=1}^{N_f},
    \end{aligned}
\label{eq:definition}
\end{equation}
where $\textbf{V}_k$ is the $k$-th video clip with $N_f$ frames degraded by arbitrary weather type, $\hat{\mathbf{V}}_k$ is the restored clip. $\phi$ is the denoising function derived from DDIM~\cite{song2020denoising}. The denoising network $\epsilon_\theta$ in the denoising function consists of a single encoder and decoder.

In the training stage, we follow the formulation of DDIM and extend the diffusion forward process to a temporal counterpart for frame-correlation modeling. The denoising network $\epsilon_\theta$ with NAFNet~\cite{chen2022simple} as the backbone is trained for noise estimation, which enforces the output of the model to approximate the proposed temporal noise model.
During the inference stage, we present a proxy task, Diff-TSC, to adapt the diffusion model to the target test data degraded by unknown weather based on the diffusion reverse process. We iteratively optimize the noise estimation of the cropped tubelets and calibrate the weight set of the model parameters for more robust weather removal.
Next, we elaborate on our diffusion-based solution for adverse weather removal in videos and our diffusion test-time adaptation, which are also displayed in Figure~\ref{fig:framework}. 


\subsection{Temporal Diffusion Process}
\label{sec:tdp}
The traditional diffusion model is trained to denoise independent noise from a perturbed image. The noise $\epsilon$ in Eq.~\eqref{eq:loss} is sampled from an i.i.d. Gaussian distribution $\epsilon \sim \mathcal{N}(0, \mathbf{I})$.
Upon training the image diffusion model and utilizing it to reverse frames from a video into the noise space individually, the significant correlation among the noise maps of these frames can be easily observed.
Rather than introducing massive parameters into the denoising network, we extend the original diffusion process to the temporal counterpart by an ARMA-based temporal noise model to efficiently preserve the correlation between frames.

\noindent\textbf{Temporal noise model. } 
ARMA (Auto-regressive Moving Average)~\cite{box2013box} is a widely used statistical model for time series data, which combines the auto-regressive (AR) part and the moving average (MA) part. Typically, ARMA can be formulated as:
\begin{equation}
    X_t = c+\varepsilon_t + \sum_{i=1}^p \varphi_i X_{t-i} + \sum_{j=1}^q \tau_j\varepsilon_{t-j},
\label{eq:arma}
\end{equation}
where $X$ is the value term, $\varepsilon$ is the error term, $c$ is a constant term, $p,q$ are the auto-regressive order and moving average order while $\varphi,\tau$ are their coefficients, respectively.
Motivated by ARMA, we develop the original Gaussian noise model $\mathcal{N}(0,\mathbf{I})$ to the temporal noise model $\mathcal{T}(0,\mathbf{I})$. Specifically, let $\{\bar{\epsilon}_i\}_{i=1}^{N_f}$ denote the noise sequence corresponding to consecutive frames of a video clip. We first sample the noise sequence $\{\bar{\epsilon}_i\}_{i=1}^{N_f}$  and the corresponding error sequence $\{\varepsilon_i\}_{i=1}^{N_f}$ from $\mathcal{N}(0,\mathbf{I})$, respectively. The constant term $c$ should be consistent with the mean of Gaussian distribution, that is, zero. To take both the past and future information into consideration, we enforce the noise of the current frame affected not only by the noise and error term of the previous frame but also those of the next frame, and correspondingly the auto-regressive order and moving average order $p,q=1$. Consequently, the expression in Eq.~\eqref{eq:arma} evolves into the convex combination of the noise term and error term:
\begin{equation}
    \bar{\epsilon}_i = (1-\varphi-\tau)\varepsilon_i + \varphi \frac{\bar{\epsilon}_{i-1}+\bar{\epsilon}_{i+1}}{2} + \tau\frac{\varepsilon_{i-1}+\varepsilon_{i+1}}{2},
\label{eq:tnm}
\end{equation}
where $\varphi,\tau$ are empirically set as $0.6,0.3$, respectively.
Algorithm~\ref{alg:alg1} further illustrates our temporal noise model, where the noise sequence of each video clip is constructed with separate preorder and postorder steps. 

\vspace{-2mm}
\begin{algorithm}
    \caption{Temporal Noise Model $\mathcal{T}(0,\mathbf{I})$}
        \label{alg:alg1}
  \KwIn{$K$ clips: $\{\mathbf{V}_{1}, \mathbf{V}_{2},  \ldots, \mathbf{V}_{K}\}$ in a batch; The coefficients $\varphi,\tau$}
    
    \For{Clip $k=1,2,...,K$ \textbf{parallelly}}{
        \For{Frame $i=1,2,...,N_f$ \textbf{parallelly}}{
            sample $\bar{\epsilon}_i, \varepsilon_i \in \mathbb{R}^{1\times c\times h\times w}$ from $\mathcal{N}(0,\mathbf{I})$\\
        }
        \For{Frame $i=2,3,...,N_f$ \textbf{sequentially}}{
            $\bar{\epsilon}_i = (1-\varphi-\tau)\varepsilon_i + \varphi \bar{\epsilon}_{i-1} + \tau\varepsilon_{i-1}$ \\
        }
        \For{Frame $i=N_f-1,...,2,1$ \textbf{sequentially}}{
            $\bar{\epsilon}_i = (1-\varphi-\tau)\varepsilon_i + \varphi \bar{\epsilon}_{i+1} + \tau\varepsilon_{i+1}$\\
        }
        $\bar{\epsilon} = normalize(\bar{\epsilon})\Longrightarrow \bar{\epsilon}\sim \mathcal{T}(0,\mathbf{I})$
    }
    \textbf{Return} $\{\{\Bar{\epsilon_i}|\Bar{\epsilon_i}\in \mathbb{R}^{1\times c\times h\times w}\}_{i=1}^{N_f}\}_{k=1}^{K}$ \\
  \end{algorithm}
To enhance the capacity of our denoising network for capturing temporal dependencies, we have specifically modified the architecture by substituting only the first and last layers of the denoising NAFNet with the 3D convolutional layer. This tailored network, denoted as $\epsilon_\theta$ is then optimized as:
\begin{equation}
 \mathcal{L}_{S} =  \|\Bar{\epsilon}-\epsilon_\theta(\sqrt{\bar{\alpha}_t}\hat{\mathbf{V}}+\sqrt{1-\bar{\alpha_t}}\Bar{\epsilon}, \mathbf{V}, t)\|_1, \\
\label{eq:loss_all}
\end{equation}
where we utilize the temporal noise model as the supervision, and $\|\cdot\|$ denotes $L_1$ norm.

\subsection{Diffusion Test-time Adaptation}
Once trained by the proposed temporal noise model, the network $\epsilon_\theta$ can generate high-quality images. This is achieved by sampling a noisy state $\mathbf{V}_T$ and then iteratively denoising it in the reverse process.
Yang \etal~\cite{yang2023video} introduced adversarial learning for weather conditions, which require weather type annotations during the training stage. Additionally, their method may fail to address out-of-distribution data, because including all real-world conditions in training, considering both weather types and image statistics, is impractical.
Contrastively, we introduce test-time adaptation and update knowledge by practice to remove arbitrary weather conditions robustly.
We design and incorporate \textbf{Diff}usion \textbf{T}ubelet \textbf{S}elf-\textbf{C}alibration (Diff-TSC), a proxy task, into the iterative denoising of the diffusion reverse process to learn the primer distribution of the frames degraded by unknown weather.

\input{tables/table1}

\vspace{-2mm}
\begin{algorithm}
    \caption{Diffusion Test-Time Adaptation to unknown weather. $\theta$ is the weight set of the trained network before adaptation, $\delta$ is the learning rate for online adaptation.}
    \label{alg:alg2}
  \KwIn{$K$ overlapped clips: $\{\mathbf{V}_{1}, \mathbf{V}_{2},  \ldots, \mathbf{V}_{K}\}$ in one video stream}
    \For{Clip $k=1,2,...,K$ \textbf{sequentially}}{
        Initialize the network $\epsilon_{\Bar{\theta}}$ with $\theta$\\
        Initialize $\hat{\mathbf{V}}_k$ by Algorithm~\ref{alg:alg1} \\
        \eIf{$k=1$}{
        \For{step $t=T,\ldots,2,1$ \textbf{sequentially}}{$\hat{\mathbf{V}}_k = ddim(\mathbf{V}_k,\hat{\mathbf{V}}_k,\epsilon_{\Bar{\theta}},t)$}
        }
        {
        $\mathbf{A}, \hat{\mathbf{A}} = TubeletCrop(\mathbf{V}_{k-1}, \hat{\mathbf{V}}_{k-1}, N_a)$\\
        
        \For{step $t=T,\ldots,2,1$ \textbf{sequentially}}{
        Compute $\mathcal{L}_S(\mathbf{A}, \hat{\mathbf{A}}, \Bar{\epsilon}, t)$ by Eq.~\eqref{eq:loss_all}\\
        $\Bar{\theta} = \Bar{\theta}-\delta\bigtriangledown_{\Bar{\theta}}\mathcal{L}_S$\\
        $\hat{\mathbf{V}}_k = ddim(\mathbf{V}_k,\hat{\mathbf{V}}_k,\epsilon_{\Bar{\theta}},t)$}
        $\hat{\mathbf{V}}_k = integrate(\hat{\mathbf{V}}_k,\hat{\mathbf{V}}_{k-1})$\\
        }

    }
    \textbf{Return} the restored clips$\{\hat{\mathbf{V}}_1,\hat{\mathbf{V}}_2,...,\hat{\mathbf{V}}_K\}$ \\

\end{algorithm}


\noindent\textbf{Diffusion Tubelet Self-Calibration. }
We design a proxy task for self-supervised learning to transfer the diffusion model to the specific weather condition. 
Specifically, for each clip from the video stream, we first initialize the weight set of $\epsilon_{\Bar{\theta}}$ with $\theta$ trained by source data. We leverage the previous clip $\mathbf{V}_{k-1}$ and its restored result $\hat{\mathbf{V}}_{k-1}$ as the pair of condition and input for the online optimization when inferring the current clip $\mathbf{V}_k$.
Rather than directly traversing all fine-grained details, we randomly crop $N_a$ tubelets from $\mathbf{V}_{k-1}$ as the primer distribution to avoid memory overhead and massive computational costs caused by high resolution.
Following the process in Section~\ref{sec:tdp}, we conduct our temporal diffusion process on the restored tubelets $\hat{\mathbf{A}}$ and feed them together with the degraded tubelets $\mathbf{A}$ into the denoising NAFNet for noise estimation. 
%
As suggested in \cite{wang2020tent}, it is necessary to iteratively find a relatively optimal point for the target distribution during test-time adaptation. Inherited from the characteristics of DDIM, such requirement can be easily achieved in the iterative diffusion reverse process, which needs more increments of inference time in the traditional CNNs and Transformers. For each timestep $t$, we first obtain the noisy tubelet $\hat{\mathbf{A}}_t$ and then optimize the objective function given in Eq.~\eqref{eq:loss_all} to update the weight set $\Bar{\theta}$. 
%
After the iterative adaptation to $\hat{\mathbf{V}}_k$, we ensemble the overlapped part of it with that of the previous clip $\hat{\mathbf{V}}_{k-1}$ to capture complementary information of pre-adaptation and post-adaptation.
We optimize over the entire network except for the last layer suggested by SHOT~\cite{liang2020we}.
The detailed procedure is described in Algorithm~\ref{alg:alg2}. 
%
We emphasize the synergy between diffusion models and test-time adaptation in accumulating frame-correlated knowledge for adverse weather removal.
By attaching the proxy task to iterative denoising, our approach also avoids excessive computational overhead caused by iterative adaptation.

%% file: tables/table1.tex
\begin{table*}[!tbp]
\centering
  \caption{\textbf{Quantitative evaluation on seen weather conditions for video adverse weather removal.} For Original Weather, these methods are trained on the weather-specific training set and tested on the weather-specific testing set. For Rain, Haze, and Snow, these methods are trained on a mixed training set and tested on the weather-specific testing set. The average performance is calculated on Rain, Haze, and Snow. PSNR and SSIM are adopted as our evaluation metrics. The best and second-best values are bold and underlined.}
  \vspace{-2mm}
    \resizebox{0.95\textwidth}{!}{%
\begin{tabular}{c|c|l|c|cc|cccccccc} 
\toprule[0.2em]
\multicolumn{2}{c|}{\multirow{2}{*}{\textbf{Methods }}}    & \multicolumn{1}{c|}{\multirow{2}{*}{\textbf{Type}}} & \multirow{2}{*}{\textbf{Source }}            & \multicolumn{10}{c}{\textbf{Datasets }}                                                                                                                                                                                                                                                                                                 \\ 
\cline{5-14}
\multicolumn{2}{c|}{}                                      & \multicolumn{1}{c|}{}                               &                                              & \multicolumn{2}{c|}{\textbf{Original Weather }} & \multicolumn{2}{c}{\textbf{Rain }}                                  & \multicolumn{2}{c}{\textbf{Haze }}                                 & \multicolumn{2}{c}{\textbf{Snow }}                                 & \multicolumn{2}{c}{\textbf{Average}}                                  \\ 
\midrule[0.2em]
\multirow{4}{*}{\textbf{Derain}}      & \textbf{PReNet}~\cite{ren2019progressive}    & Image                                      & CVPR'19                                      & 27.06 & 0.9077                                  & 26.80                           & 0.8814                            & 17.64                           & 0.8030                           & 28.57                           & 0.9401                           & 24.34~                           & 0.8748~                            \\
                                      & \textbf{SLDNet}~\cite{yang2020self}    & Video                                      & CVPR'20                                      & 20.31 & 0.6272                                  & 21.24                           & 0.7129                            & 16.21                           & 0.7561                           & 22.01                           & 0.8550                           & 19.82~                           & 0.7747~                            \\
                                      & \textbf{S2VD}~\cite{yue2021semi}      & Video                                      & CVPR'21                                      & 24.09 & 0.7944                                  & 28.39                           & 0.9006                            & 19.65                           & 0.8607                           & 26.23                           & 0.9190                           & 24.76~                           & 0.8934~                            \\
                                      & \textbf{RDD-Net}~\cite{wang2022rethinking}   & Video                                      & ECCV'22                                      & 31.82 & 0.9423                                  & 30.34                           & 0.9300~                           & 18.36                           & 0.8432                           & 30.40~                          & 0.9560~                          & 26.37~                           & 0.9097~                            \\ 
\hline
\multirow{4}{*}{\textbf{Dehaze }}     & \textbf{GDN}~\cite{liu2019griddehazenet}       & Image                                      & ICCV'19                                      & 19.69 & 0.8545                                  & 29.96                           & 0.9370~                           & 19.01                           & 0.8805                           & 31.02                           & 0.9518                           & 26.66~                           & 0.9231~                            \\
                                      & \textbf{MSBDN}~\cite{dong2020multi}     & Image                                      & CVPR'20                                      & 22.01 & 0.8759                                  & 26.70                           & 0.9146~                           & 22.24                           & 0.9047                           & 27.07                           & 0.9340                           & 25.34~                           & 0.9178~                            \\
                                      & \textbf{VDHNet}~\cite{ren2018deep}    & Video                                      & TIP'19                                       & 16.64 & 0.8133                                  & 29.87                           & 0.9272~                           & 16.85                           & 0.8214                           & 29.53                           & 0.9395                           & 25.42~                           & 0.8960~                            \\
                                      & \textbf{PM-Net}~\cite{liu2022phase}    & Video                                      & MM'22                                        & 23.83 & 0.8950~                                 & 25.79                           & 0.8880~                           & 23.57                           & 0.9143                           & 18.71                           & 0.7881                           & 22.69~                           & 0.8635~                            \\ 
\hline
\multirow{4}{*}{\textbf{Desnow }}     & \textbf{DesnowNet}~\cite{liu2018desnownet} & Image                                     & TIP'18                                       & 28.30 & 0.9530                                  & 25.19                           & 0.8786                            & 16.43                           & 0.7902                           & 27.56                           & 0.9181                           & 23.06~                           & 0.8623~                            \\
                                      & \textbf{DDMSNET}~\cite{zhang2021deep}   & Image                                      & TIP'21                                       & 32.55 & 0.9613                                  & 29.01                           & 0.9188~                           & 19.50                           & 0.8615                           & \textbf{32.43}                           & \textbf{0.9694}                          & 26.98~                           & 0.9166~                            \\
                                      & \textbf{HDCW-Net}~\cite{chen2021all}  & Image                                      & ICCV'21                                      & 31.77 & 0.9542                                  & 28.10~                          & 0.9055~                           & 17.36                           & 0.7921                           & 31.05                           & 0.9482~                          & 25.50~                           & 0.8819~                            \\
                                      & \textbf{SMGARN}~\cite{cheng2022snow}    & Image                                      & \textcolor[rgb]{0.141,0.161,0.184}{TCSVT‘22} & 33.24 & 0.9721                                  & 27.78                           & 0.9100~                           & 17.85                           & 0.8075                           & \underline{32.34}                           & \underline{0.9668}                           & 25.99~                           & 0.8948~                            \\ 
\hline
\multirow{5}{*}{\textbf{Restoration}} & \textbf{MPRNet}~\cite{zamir2021multi}    & Image                                      & CVPR’21                                      & — —    & — —                                      & 28.22~                          & 0.9165~                           & 20.25                           & 0.8934                           & 30.95                           & 0.9482~                          & 26.47~                           & 0.9194~                            \\
                                      & \textbf{EDVR}~\cite{wang2019edvr}      & Video                                      & CVPR'19                                      & — —    & — —                                      & 31.10~                          & 0.9371~                           & 19.67                           & 0.8724                           & 30.27                           & 0.9440~                          & 27.01~                           & 0.9178~                            \\
                                      & \textbf{RVRT}~\cite{liang2022recurrent}      & Video                                      & NIPS'22                                      & — —    & — —                                      & 30.11~                          & 0.9132~                           & 21.16                           & 0.8949                           & 26.78                           & 0.8834                           & 26.02~                           & 0.8972~                            \\
                                      & \textbf{RTA}~\cite{zhou2022revisiting}       & Video                                      & CVPR'22                                      & — —    & — —                                      & 30.12~                          & 0.9186~                           & 20.75                           & 0.8915                           & 29.79                           & 0.9367                           & 26.89~                           & 0.9156~                            \\ 
\hline
\multirow{8}{*}{\textbf{Multi-Weather}} & \textbf{All-in-one}~\cite{li2020all}    & Image                                      & CVPR‘20                                      & — —    & — —                                      & 26.62                           & 0.8948~                           & 20.88                           & 0.9010                           & 30.09                          & 0.9431                           & 25.86~                           & 0.9130~                            \\
                                      & \textbf{UVRNet}~\cite{kulkarni2022unified}      & Image                                      & TMM'22                                      & — —    & — —                                      & 22.31                           & 0.7678~                           & 20.82                           & 0.8575                           & 24.71                           & 0.8873                           & 22.61~                           & 0.8375~                            \\
                                      & \textbf{TransWeather}~\cite{valanarasu2022transweather}      & Image                                      & CVPR'22                                      & — —    & — —                                      & 26.82                           & 0.9118~                           & 22.17                           & 0.9025                           & 28.87                           & 0.9313                           & 25.95~                           & 0.9152~                            \\
                                      & \textbf{TKL}~\cite{chen2022learning}       & Image                                      & CVPR'22                                      & — —    & — —                                      & 26.73                           & 0.8935~                           & 22.08                           & 0.9044                           & 31.35                           & 0.9515                           & 26.72~                           & 0.9165~                            \\
                                      & \textbf{WeatherDiffusion}~\cite{ozdenizci2023restoring}       & Image                                      & TPAMI'23                                      & — —    & — —                                      & 25.86                           & 0.9125                          & 20.10                           & 0.8442                           & 26.40                           & 0.9113                           & 24.12~                           & 0.8893~                            \\

                                      & \textbf{WGWS-Net}~\cite{zhu2023learning}       & Image                                      & CVPR'23                                      & — —    & — —                                      & 29,64                           & 0.9310                           & 17.71                           & 0.8113                         & 31.58                           & 0.9528                           & 26.31                           & 0.9265                            \\
                                      & \textbf{ViWS-Net}~\cite{yang2023video}       & Video                                      & ICCV'23                                          & — —    & — —                                      & \underline{31.52} & \underline{0.9433} & \underline{24.51} & \textbf{0.9187} & 31.49 & 0.9562 & \underline{29.17} & \underline{0.9394}  \\
                                      & \cellcolor{mygray}\textbf{Ours }       & \cellcolor{mygray}Video                                      & \cellcolor{mygray}— —                                      & \cellcolor{mygray}— —    & \cellcolor{mygray}— —                                      & \cellcolor{mygray}\textbf{32.43}~                          & \cellcolor{mygray}\textbf{0.9573}~                           & \cellcolor{mygray}\textbf{24.56}                           & \cellcolor{mygray}\underline{0.9148}                           & \cellcolor{mygray}31.86                           & \cellcolor{mygray}0.9640                           & \cellcolor{mygray}\textbf{29.63~}                          & \cellcolor{mygray}\textbf{0.9453~}                            \\

\bottomrule[0.2em]
\end{tabular}
    }
\label{tab:main}
\vspace{-2mm}
\end{table*}

%% file: section4-experiments.tex
\vspace{-1mm}
\section{Experiments}

\input{tables/table2}

\begin{figure*}
\centering
\includegraphics[width=\textwidth]{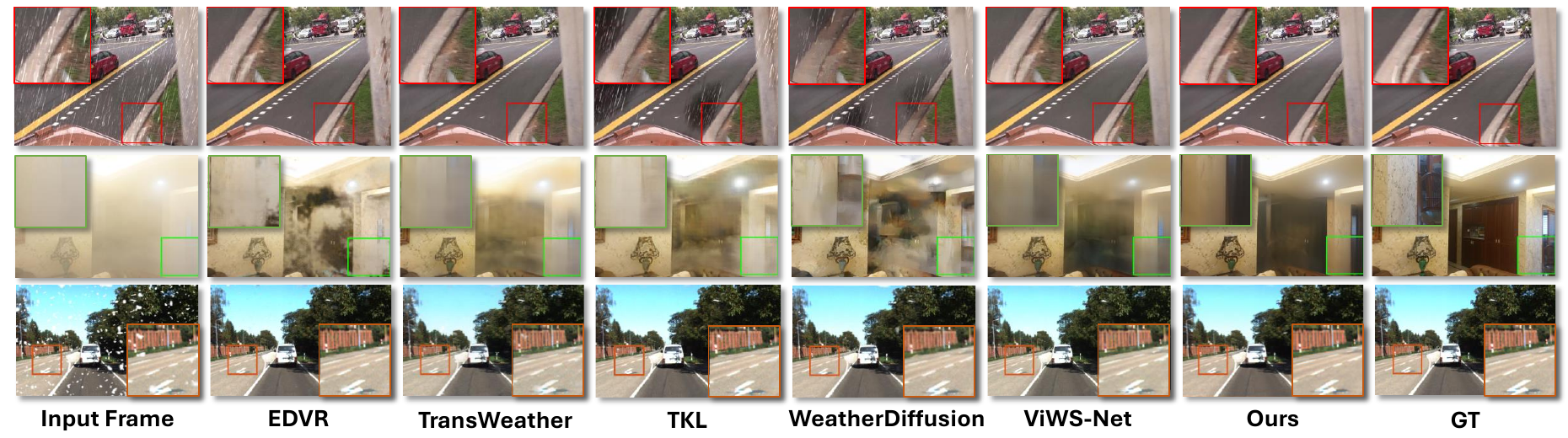}
\vspace{-5mm}
\caption{\textbf{Qualitative Comparison of seen weather conditions between our approach and state-of-the-art algorithms.} The competitive algorithms are selected to present the results on the example frames degraded by rain, haze, snow, respectively. The color box indicates the comparison of details. Please zoom in on the images for improved visualization.}
\label{fig:vis_comparison}
\vspace{-5mm}
\end{figure*}

\subsection{Datasets}
Various video adverse weather datasets are used in our experiments. Following the setting of \cite{yang2023video}, we adopt RainMotion~\cite{wang2022rethinking}, REVIDE~\cite{zhang2021learning} and KITTI-snow~\cite{yang2023video} as seen weather. For the training stage, we merge the training set of the three datasets as the mixed set to learn a generic model. For the testing stage, we evaluate our model on three testing sets, respectively.
Additionally, we evaluate our approach on two datasets VRDS~\cite{wu2023mask} and RVSD~\cite{chen2023snow} to validate the generalization of our Diff-TTA on unseen weather conditions. VRDS is a synthesized video dataset of joint rain streaks and raindrops with a total of 102 videos, while RVSD is a realistic video desnowing dataset with a total of 110 videos containing both snow and fog achieved by the rendering engine. Due to the large quantity of the two datasets, we randomly sample 8 videos as the test set, each of which has 20 frames, respectively. We also collect several real-world videos degraded by diverse weather conditions to prove its effectiveness on real-world applications.

\subsection{Implementation}
\noindent \textbf{Training Details.} 
The proposed framework was trained on NVIDIA RTX 4090 GPUs and implemented on the Pytorch platform. We adopted the Lion optimizer~\cite{chen2023symbolic} ($\beta_1=0.9,\beta_2=0.99$) with the initial learning rate of $2 \times 10^{-5}$ decayed to $2 \times 10^{-7}$ by the Cosine scheduler. We randomly crop the video frames to 256$\times$256 for training. Our framework is empirically trained for \textit{600k} iterations in an end-to-end way with batch size of 4. Each batch consists of 4 video clips with 5 frames per clip.

\noindent \textbf{Method-specific Details. }
During the training, we used the diffusion process linearly increasing from $\beta_1=0.0001$ to $\beta_T=0.02$ with $T = 1000$ timesteps.
During the testing, the inference step is reduced to 25 by deterministic implicit sampling. We utilize the same optimizer with a smaller learning rate of $1\times10^{-6}$ for the test-time adaptation. 
The number of cropped tubelets $N_a$ is empirically set to 5 while the crop size is 128$\times$128. 
Despite the incorporated adaptation, our Diff-TTA is particularly 90$\times$ more efficient than WeatherDiffusion. To process a video clip of 5 frames with the patch size of 256$\times$256, the average run time of Diff-TTA is 6.01s while that of WeatherDiffusion is 542.76s.

\subsection{Performance Evaluation}
\noindent \textbf{Comparison methods. }
As shown in Table~\ref{tab:main}, we compared our proposed method against five kinds of state-of-the-art methods on the three seen weather conditions as \cite{yang2023video} does, \ie, \textit{derain, dehaze, desnow, restoration, all-in-one adverse weather removal}. 
For all-in-one adverse weather removal, we compared ours with the six representative single-image methods All-in-one~\cite{li2020all}, UVRNet~\cite{kulkarni2022unified}, TransWeather~\cite{valanarasu2022transweather}, TKL~\cite{chen2022learning}, WeatherDiffusion~\cite{ozdenizci2023restoring}, WGWS-Net~\cite{zhu2023learning}, and the video-level method ViWS-Net~\cite{yang2023video}.

\input{tables/table3}

\begin{figure*}
\centering
\includegraphics[width=\textwidth]{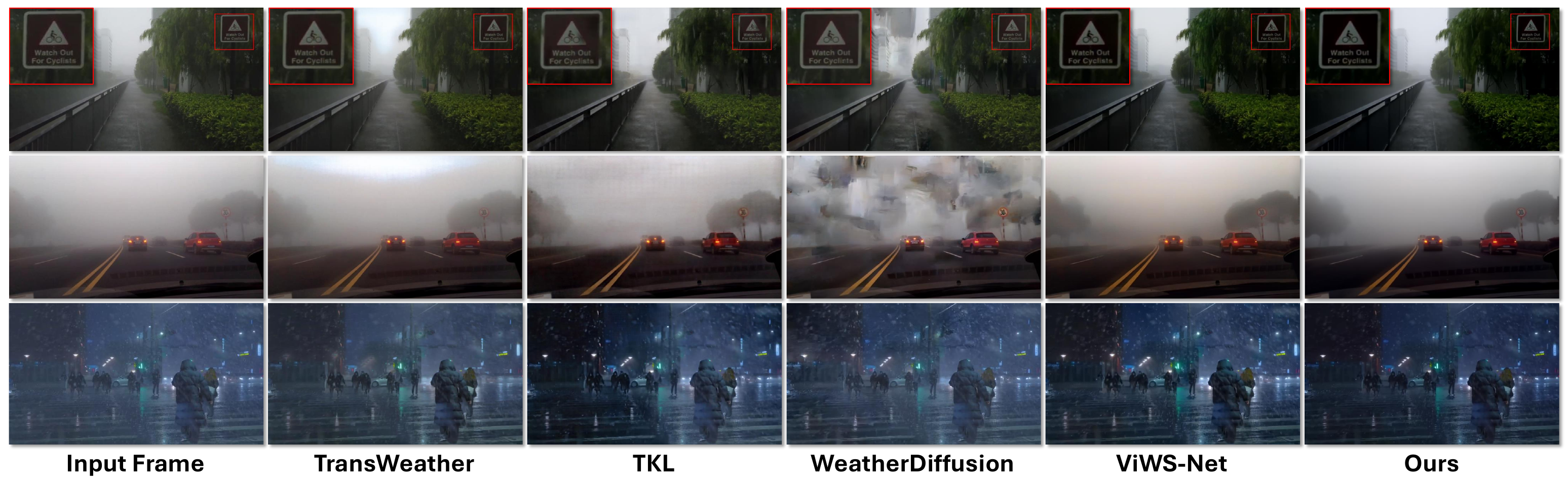}
\vspace{-6mm}
\caption{\textbf{Visual comparison of all-in-one adverse weather removal methods on the selected real-world video sequences degraded by rain, haze, snow.} Apparently, our network can more effectively remove rain streaks, haze, and snowflakes of input video frames than state-of-the-art methods. Please zoom in on the images for improved visualization.}
\label{fig:realworld}
\vspace{-4mm}
\end{figure*}

\begin{figure}
\centering
\includegraphics[width=0.95\columnwidth]{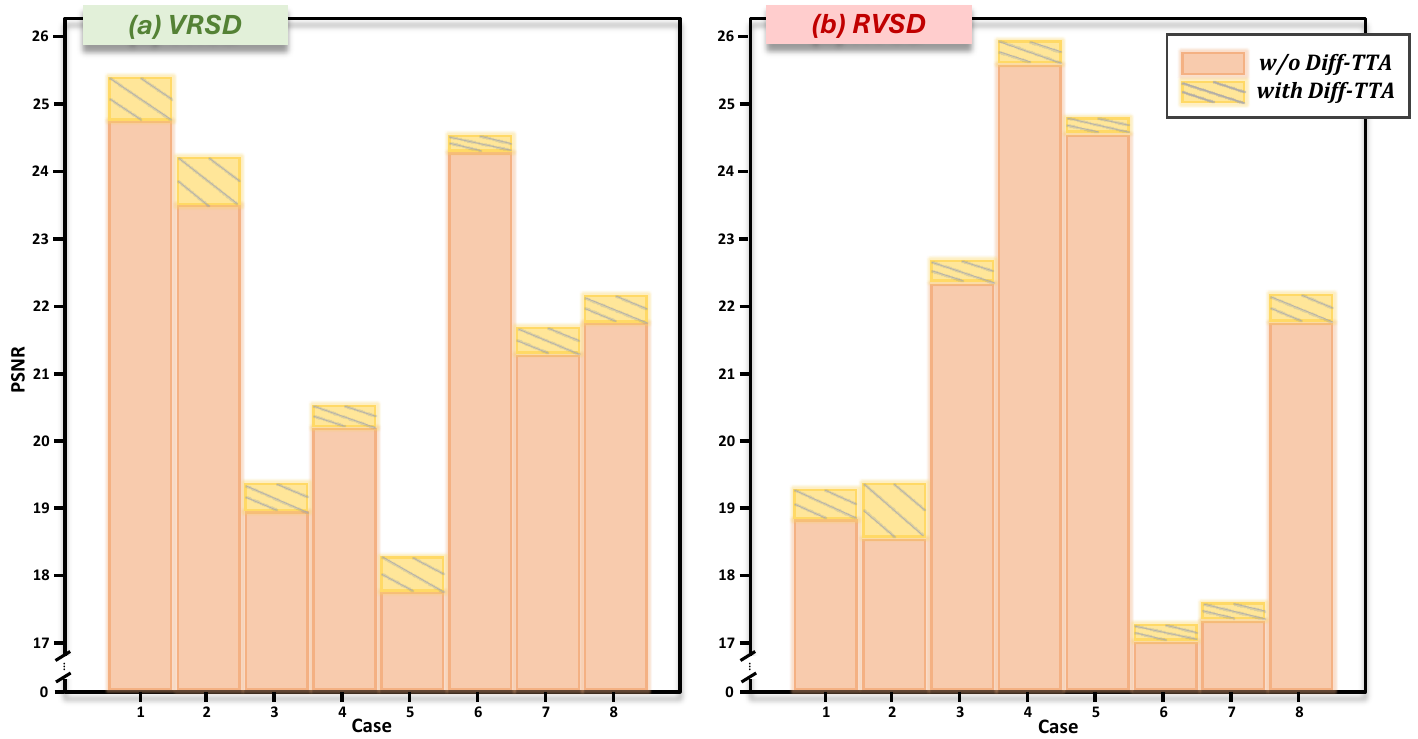}
\vspace{-2mm}
\caption{Detailed results on each case of VRDS and RVSD by our proposed Diff-TTA. The proposed Diff-TTA brings consistent improvement across various cases. }
\label{fig:unseen_case}
\vspace{-5mm}
\end{figure}

\noindent \textbf{Quantitative Comparison. }
\textit{For quantitative evaluation of the restored results, we apply the peak signal-to-noise ratio (PSNR) and the structural similarity (SSIM) as the metrics.}
Following the previous experimental setting, only the results of the model trained on the mixed training set are reported for all-in-one adverse weather removal models.
To ensure the fairness of comparisons, we retrain each compared model implemented by the official codes based on our training dataset and report the best result.
Observed from Table~\ref{tab:main}, our method achieves the best average performance across rain, haze and snow conditions among all kinds of methods by a considerable margin of 0.46, 0.0059 in PSNR, SSIM, respectively, than the second-best method ViWS-Net~\cite{yang2023video}. Though some single-weather methods like DDMSNET~\cite{zhang2021deep} and SMGARN~\cite{cheng2022snow} achieve impressive results in their own weather condition, they frequently fail to remove other weather conditions, such as haze. Our method significantly surpasses other all-in-one adverse weather methods in rain and snow conditions, and obtains competitive results in haze conditions. It is worth noting that the generalizable ability is critical for such methods, which can be observed from the average performance. To further validate their generalization, we evaluate all-in-on adverse weather removal methods on unseen weather conditions, \ie, rain streak+raindrop and snow+fog, in Table~\ref{tab:unseen}.
Our Diff-TTA exhibits remarkable superiority over state-of-the-art methods when confronted with these unseen weather conditions.

\noindent \textbf{Visual comparisons on synthetic videos.}
To vividly demonstrate the effectiveness of our approach, Figure~\ref{fig:vis_comparison} shows the qualitative comparison under the rain, haze, and snow conditions between our method and five state-of-the-art methods 
Our method demonstrates promising results in terms of visual quality across various weather types. Specifically, when it comes to rain and snow scenarios, our approach significantly reduces the presence of rain streaks and snow particles, outperforming other methods. In hazy scenarios, our method effectively removes residual haze and preserves the clean background with impressive results. 
\textit{More visual results of seen and unseen weather are displayed in the supplementary files.}

\noindent \textbf{Visual comparisons on real-world videos.}
To evaluate the universality of our Diff-TTA on real-world applications, we collect several real-world degraded videos by rain, haze, snow from Youtube website, and further compare our approach against other all-in-one adverse weather removal methods. Figure~\ref{fig:realworld} displays the visual results produced by our network and the compared methods on real-world video frames, clearly showcasing our exceptional performance. 

\noindent \textbf{Ablation Study.}
We evaluate the effectiveness of each component including diffusion process, temporal noise model, and Diffusion test-time adaptation (Diff-TTA) as described in Table~\ref{tab:ablation}. We present the results trained on the mixed training set and tested on three weather-specific testing sets (rain, haze, snow) in view of PSNR and SSIM. The baseline model M1, \ie, the vanilla NAFNet, achieves the average performance on the three adverse weather datasets of 26.83, 0.9264 in PSNR, SSIM, respectively. M2 is equipped with diffusion process and trained by noise estimation, which advances M1 on the average performance by 1.69, 0.0078 of PSNR, SSIM, respectively. M3 introduces the proposed temporal noise model by replacing the vanilla noise model, and brings about an improvement of 0.63, 0.0083 in average PSNR, SSIM, respectively. Finally, our full model employs diffusion test-time adaptation (Diff-TTA) and gains a critical increase of 0.48, 0.0028 in PSNR, SSIM, respectively, compared to M3.
Figure~\ref{fig:unseen_case} further ablates our iterative adaptation on each case of the two unseen datasets, demonstrating consistent improvements.

%% file: tables/table2.tex
\begin{table}[t]
  \centering
  \caption{\textbf{Quantitative evaluation on unseen weather conditions for video adverse weather removal.} The best and second-best performances are marked in bold and underlined, respectively.}
  \resizebox{1\linewidth}{!}{
    \begin{tabular}{l|cc|cc}
    \toprule[0.2em]
    \multirow{2}[2]{*}{\textbf{Method}} & \multicolumn{2}{c|}{\textbf{VRDS}~\cite{wu2023mask}} & \multicolumn{2}{c}{\textbf{RVSD}~\cite{chen2023snow}} \\
         &  \multicolumn{1}{c}{PSNR $\uparrow$} & \multicolumn{1}{c|}{SSIM $\uparrow$} & \multicolumn{1}{c}{PSNR $\uparrow$} & \multicolumn{1}{c}{SSIM $\uparrow$} \\
    \midrule[0.2em]
    All-in-one~\cite{li2020all}  & \multicolumn{1}{c}{20.44} & \multicolumn{1}{c|}{0.5944} & \multicolumn{1}{c}{19.79} & \multicolumn{1}{c}{0.7509} \\
    TransWeather~\cite{valanarasu2022transweather} & \multicolumn{1}{c}{21.36} & \multicolumn{1}{c|}{\underline{0.7136}} & \multicolumn{1}{c}{\underline{20.25}} & \multicolumn{1}{c}{0.7514} \\
    TKL~\cite{chen2022learning} & \multicolumn{1}{c}{20.49} & \multicolumn{1}{c|}{0.7003} & \multicolumn{1}{c}{19.71} & \multicolumn{1}{c}{0.7370} \\
    WeatherDiffusion~\cite{ozdenizci2023restoring} & \multicolumn{1}{c}{20.73} & \multicolumn{1}{c|}{0.6943} & \multicolumn{1}{c}{18.08} & \multicolumn{1}{c}{0.6588} \\
    ViWS-Net~\cite{yang2023video} & \multicolumn{1}{c}{\underline{21.57}} & \multicolumn{1}{c|}{0.7094} & \multicolumn{1}{c}{19.83} & \multicolumn{1}{c}{\underline{0.7590}} \\

    \midrule[0.2em]

    \rowcolor{mygray}Diff-TTA (ours) &   \multicolumn{1}{c}{\textbf{22.57}}    &       \multicolumn{1}{c|}{\textbf{0.7281}}    & \multicolumn{1}{c}{\textbf{22.35}} & \multicolumn{1}{c}{\textbf{0.7719}} \\
    \bottomrule[0.2em]
    \end{tabular}%
    }
  \label{tab:unseen}%
  \vspace{-4mm}
\end{table}%

%% file: tables/table3.tex
\begin{table*}[!tbp]
\centering
  \caption{\textbf{Ablation study of each critical component in the proposed framework on seen weather conditions.} The top values are marked in bold font. ``Diff-TTA'' denote Diffusion Test-time Adaptation.}
\vspace{-2mm}
    \resizebox{0.9\textwidth}{!}{%
\begin{tabular}{c|ccc|cc|cc|cc|cc} 
\toprule[0.2em]
\multirow{2}*{\textbf{Combination}} & \multicolumn{3}{c|}{\textbf{Component}}                                      & \multicolumn{8}{c}{\textbf{Datasets}}                                                                            \\ 
\cline{2-12}
                       & \textbf{Diffusion Process} & \textbf{Temporal Noise} & \textbf{Diff-TTA} & \multicolumn{2}{c|}{Rain} & \multicolumn{2}{c|}{Haze} & \multicolumn{2}{c|}{Snow} & \multicolumn{2}{c}{Average}  \\ 
\midrule[0.2em]
M1                              & \textbf{-}                         & \textbf{-}                     & \textbf{-}                     & 29.07 & 0.9514            & 22.64 & 0.8930            & 28.79 & 0.9350            & 26.83 & 0.9264               \\
M2                                    & \checkmark                         & \textbf{-}                     & \textbf{-}                     & 31.55 & 0.9466            & 23.58 & 0.9041            & 30.48 & 0.9520            & 28.52 & 0.9342               \\

M3                                    & \checkmark                         &\checkmark                     & \textbf{-}                     & 32.10 & 0.9530            & 24.31 & 0.9124            & 31.04 & 0.9621            & 29.15 & 0.9425               \\

\textbf{Ours}                         &\checkmark                         & \checkmark                     & \checkmark                     &    \textbf{32.43}   &    \textbf{0.9573}               &    \textbf{24.56}   &    \textbf{0.9148}               &    \textbf{31.86}   &    \textbf{0.9640}               &   \textbf{29.63}    &    \textbf{0.9453}                  \\
\bottomrule[0.2em]

\end{tabular}
    }
\label{tab:ablation}
\vspace{-2mm}
\end{table*}

%% file: section5-conclusion.tex

\section{Conclusion}

This paper presents Diff-TTA, a diffusion-based test-time adaptation method capable of restoring videos degraded by various, even unknown, weather conditions. The proposed approach involves constructing a diffusion-based network and utilizing a novel temporal noise model to establish correlations among video frames. More importantly, a proxy task Diff-TSC is incorporated into the diffusion reverse process to iteratively update the deep model. This allows for the efficient adaptation of vision systems in real-world scenarios without modifying the source training process, making it easily applicable to a wide range of off-the-shelf diffusion-based models for enhancing weather removal. Experimental results on both synthetic and real-world data, which encompass diverse weather conditions, confirm the effectiveness and generalization capability of the proposed approach. We believe Diff-TTA can be a compelling baseline that shifts the focus of the community from the complicated training process to online test-time adaptation.

%% file: X_suppl.tex
\clearpage
\setcounter{page}{1}
\maketitlesupplementary

This is a supplementary material for Genuine Knowledge from Practice: Diffusion Test-Time Adaptation for Video Adverse Weather Removal.

We provide the following materials in this manuscript:
\begin{itemize}
    \item Sec.~\ref{sec:supp_details} more details of our designed method.
    \item Sec.~\ref{sec:supp_compute} computational costs comparison.
    \item Sec.~\ref{sec:supp_visual} visualization results
    \item Sec.~\ref{sec:supp_future} future work
\end{itemize}

\section{More Details}
\label{sec:supp_details}

\begin{figure}
\centering
\includegraphics[width=\columnwidth]{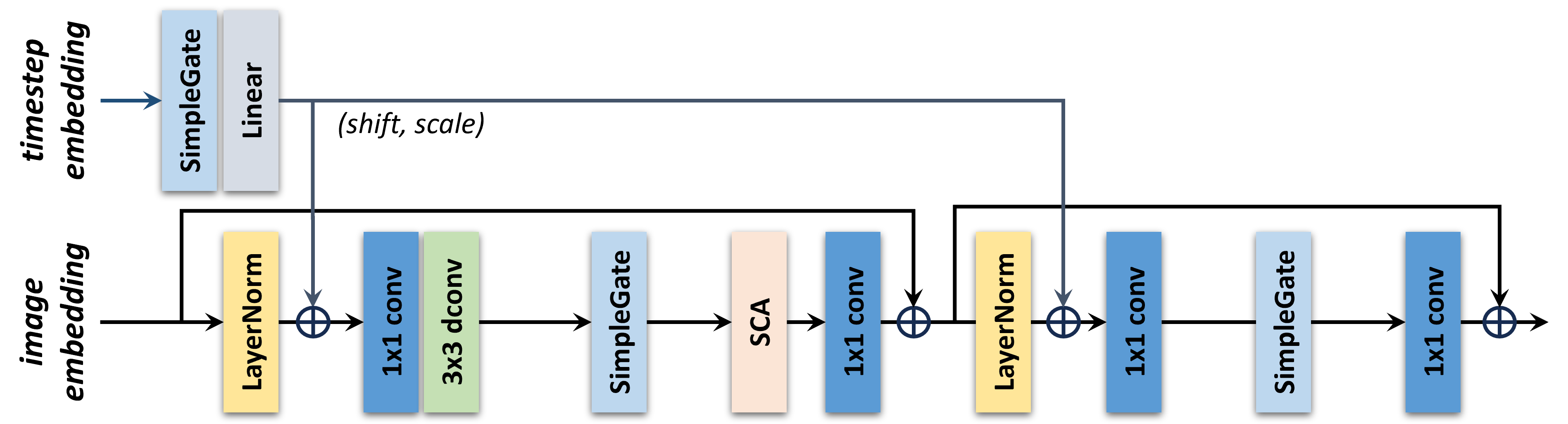}
\caption{Detailed structure of the Denoising NAFNet's block.}
\label{fig:supp_nafblock}
\end{figure}

In our proposed method, we adopt NAFNet~\cite{chen2022simple} as the backbone of the denoising network. NAFNet is a very simple but efficient baseline for image restoration task. We adapt it to video restoration tasks based on its original setting.
Specifically, it is enlarged by increasing the width to 64, the number of blocks of NAFNet to 44. The additional MLP layer comprised of SimpleGate and a linear layer is incorporated into each block for time embedding as shown in Figure~\ref{fig:supp_nafblock}.

\section{Computational Costs Comparison}
\label{sec:supp_compute}
In this section, we showcase our speed advantages against diffusion-based restoration methods. We conduct all inference stages on an RTX4090 GPU to ensure a fair comparison. The results of FLOPs and run-time on a video clip of 5 frames with 256$\times$256 patch size are displayed in Table~\ref{tab:complexity}. This demonstrates that our approach achieves superior model complexity and run-time performance after incorporating the adaptation mechanism while obtaining promising restoration performance.

\section{More Visualization Results}
\label{sec:supp_visual}
We present more visual comparisons against state-of-the-art methods on synthetic and real datasets to demonstrate the excellent visual performance and generalization ability of Diff-TTA in removing arbitrary adverse weather conditions. The restored results of consecutive frames with diverse weather conditions are also displayed in the supplementary videos.

\begin{figure*}
\centering
\includegraphics[width=\textwidth]{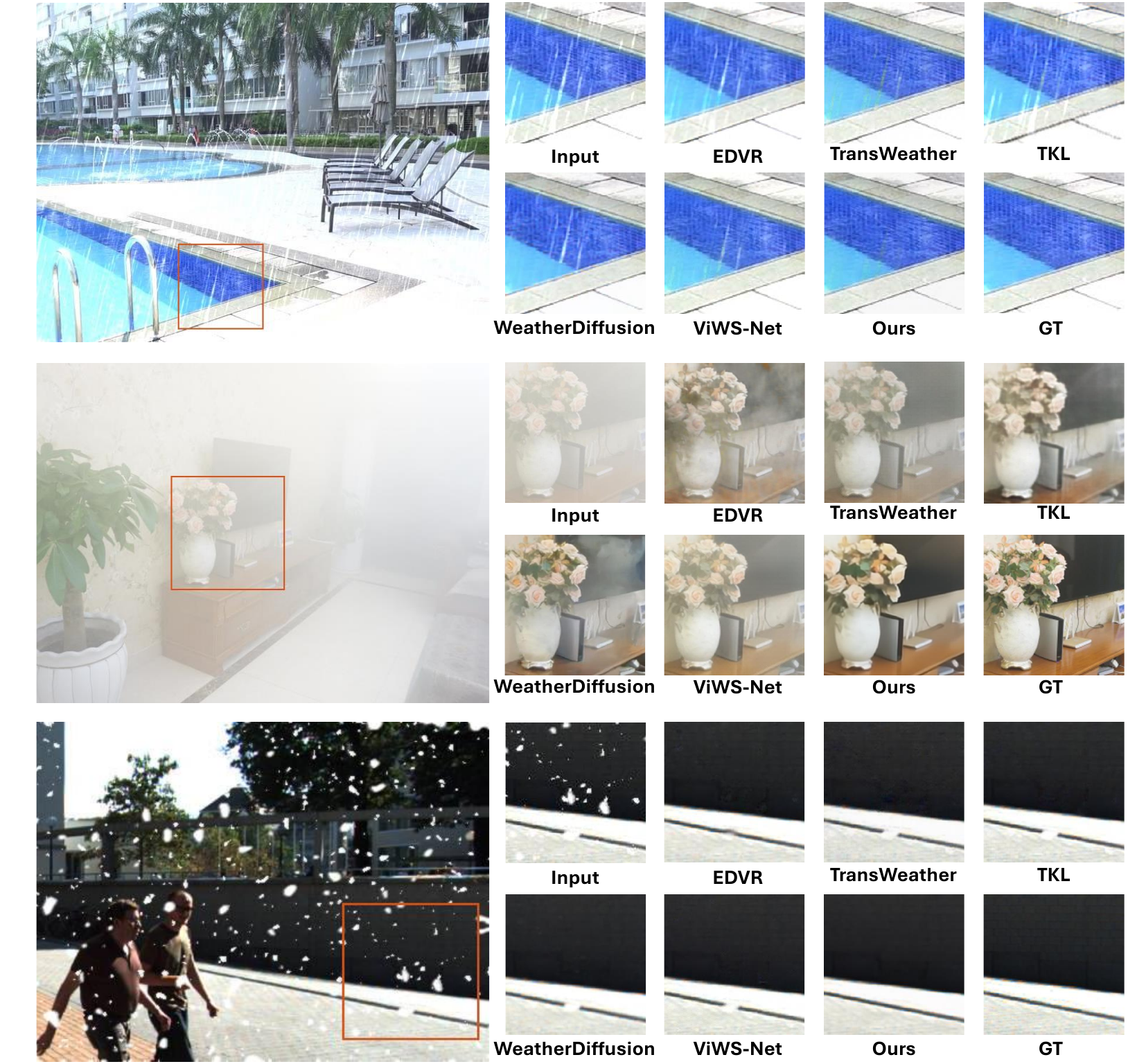}
\caption{More visual comparisons on seen weather conditions.}
\label{fig:supp_seen}
\end{figure*}

\noindent\textbf{Visualization comparison on seen datasets.}
We provide more qualitative comparisons between our Diff-TTA and SOTA methods. The results are shown in Figure~\ref{fig:supp_seen}. Our Diff-TTA achieves the best visual quality by removing adverse components and recovering more background details.

\begin{figure*}
\centering
\includegraphics[width=\textwidth]{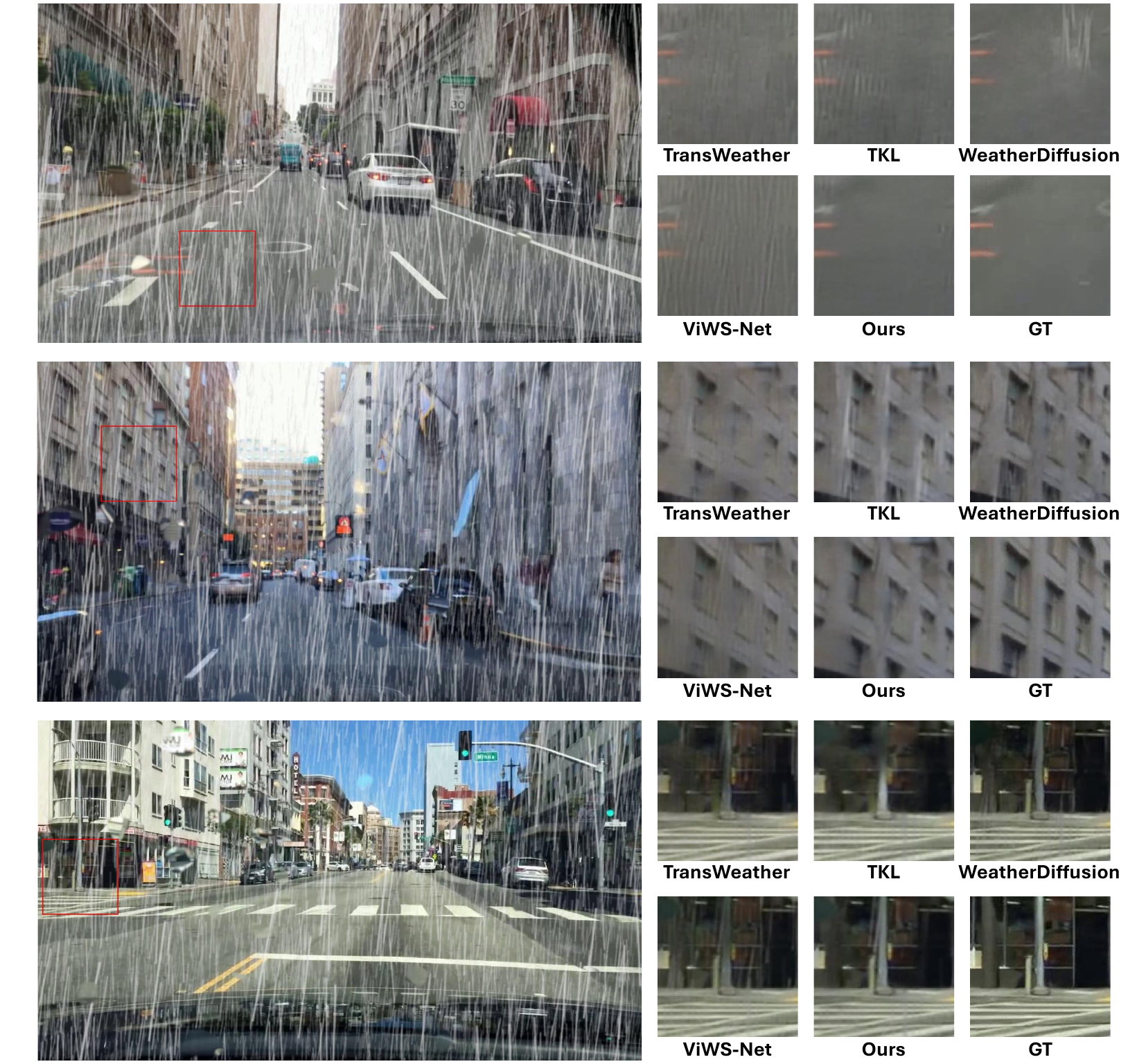}
\caption{More visual comparisons on unseen weather conditions.}
\label{fig:supp_unseen}
\end{figure*}

\begin{figure*}
\centering
\includegraphics[width=\textwidth]{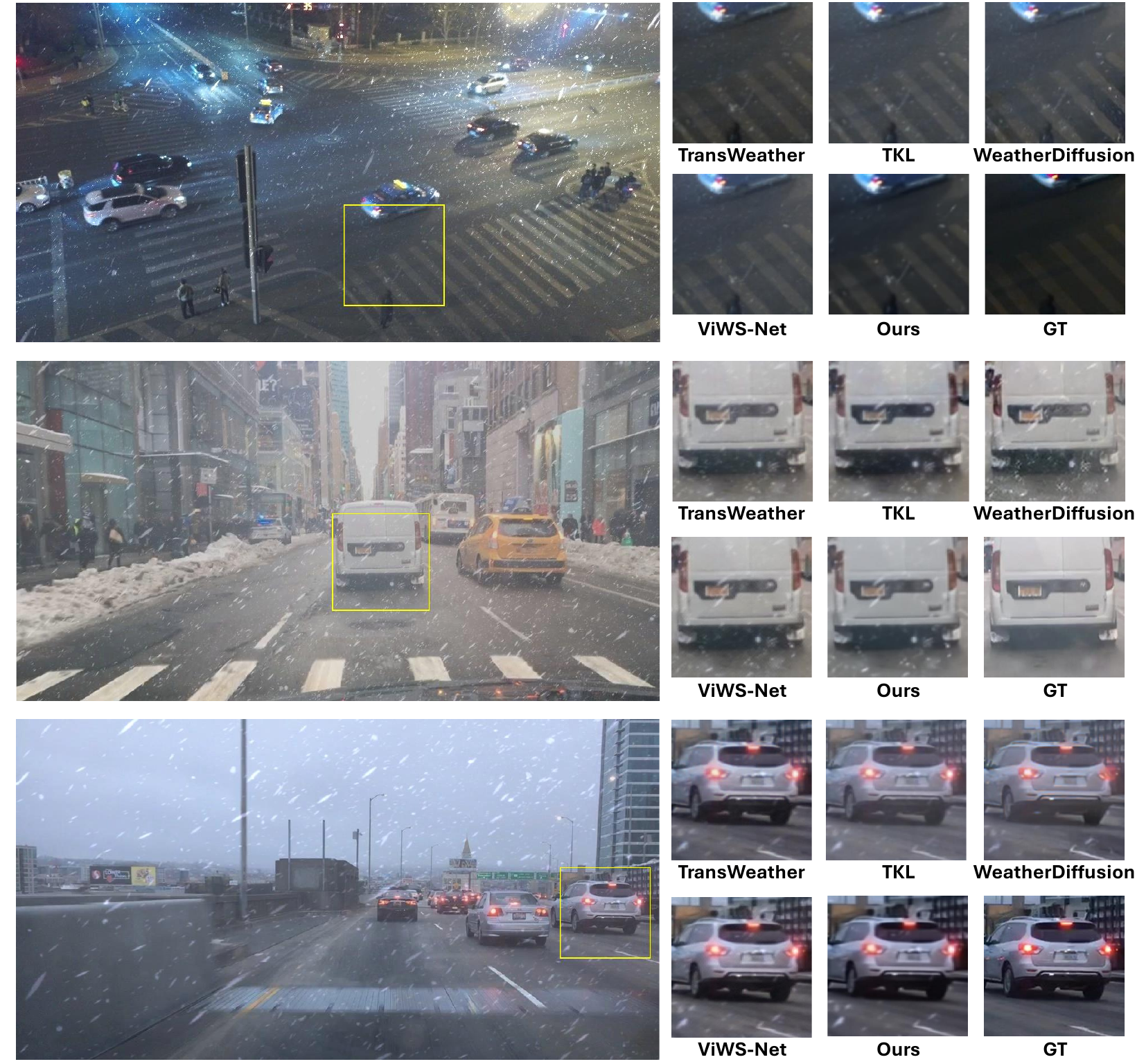}
\caption{More visual comparisons on unseen weather conditions.}
\label{fig:supp_unseen2}
\end{figure*}

\noindent\textbf{Visualization comparison on unseen datasets.}
We provide qualitative comparisons between our Diff-TTA and SOTA methods. The results are shown in Figure~\ref{fig:supp_unseen}, \ref{fig:supp_unseen2}. Our Diff-TTA achieves the best visual quality by removing unknown adverse components and recovering more background details.

\begin{figure*}
\centering
\includegraphics[width=0.8\textwidth]{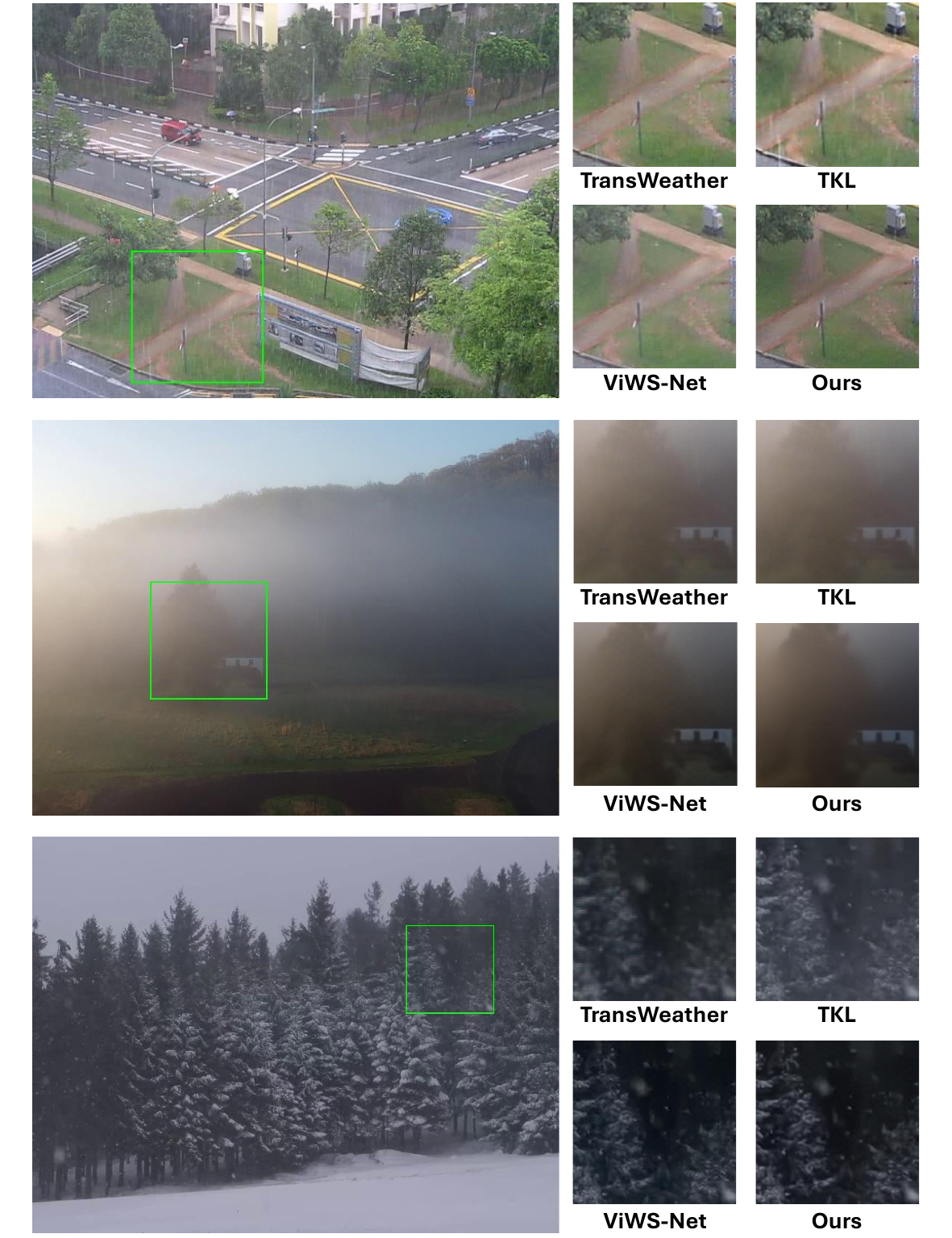}
\caption{More visual comparisons on real-world scenarios.}
\label{fig:supp_real}
\end{figure*}

\noindent\textbf{Visualization comparison on real-world data.}
We provide more qualitative comparisons between our Diff-TTA and SOTA methods. The results are shown in Figure~\ref{fig:supp_real}. Our Diff-TTA achieves the best visual quality by removing unknown adverse components and recovering more background details, further validating its generalization.

\begin{table}[t]
  \centering
  \caption{Comparison of FLOPs and Run-time on a video clip of 5 frames with 256$\times$256 patch size.}
  \resizebox{1\linewidth}{!}{
    \begin{tabular}{lcccc}
    \toprule[0.2em]
    \rowcolor{mygray}\textbf{Method} & IR-SDE~\cite{luo2023image} & Refusion~\cite{luo2023refusion} & WeatherDiffusion~\cite{ozdenizci2023restoring} & Diff-TTA(ours) \\
    \hline
    FLOPs/G     & 1896.67$\times$100 steps & 372.38$\times$100 steps & 1242.01$\times$1000 steps & 385.14$\times$25 steps \\
    Run-time/s  & 23.42 & 12.55 & 542.76 & 6.01 \\
    \bottomrule[0.2em]
    \end{tabular}%
    }
  \label{tab:complexity}%
  \vspace{-4mm}
\end{table}%

\section{Future Work}
\label{sec:supp_future}
In the future, we plan further to accelerate the inference speed of Diff-TTA for real-time applications. Additionally, we will apply test-time adaptation to other restoration tasks.

